\documentclass[sigconf]{acmart}
\AtBeginDocument{%
  }

\copyrightyear{2026}
\acmYear{2026}
\setcopyright{cc}
\setcctype{by}
\acmConference[DAC '26]{63rd ACM/IEEE Design Automation Conference}{July 26--29, 2026}{Long Beach, CA, USA}
\acmBooktitle{63rd ACM/IEEE Design Automation Conference (DAC '26), July 26--29, 2026, Long Beach, CA, USA}
\acmDOI{10.1145/3770743.3804360}
\acmISBN{979-8-4007-2254-7/2026/07}

\usepackage{color}
\usepackage{listings}
\usepackage{framed}
\usepackage[dvipsnames,table]{xcolor}
\usepackage{enumitem}
\usepackage{booktabs}
\usepackage{xspace}
\usepackage{multirow}
\usepackage{threeparttable}
\usepackage{caption}
\usepackage{subcaption}
\usepackage{capt-of}
\usepackage{graphicx}
\newcommand{\ie}{\emph{i.e.,}\xspace}

\newcommand{\eg}{\emph{e.g.,}\xspace}

\newcommand{\Improve}[1]{\textcolor{OliveGreen}{($\downarrow$#1)}}
\newcommand{\Decrease}[1]{\textcolor{Maroon}{($\uparrow$#1)}}
\newcommand{\Original}[1]{\textcolor{Maroon}{(#1)}}

\newcommand{\ThroughputImprove}[1]{\textcolor{OliveGreen}{($\uparrow$#1)}}

\newcommand{\DeInferName}{{DeInfer}}
\newcommand{\DeInfer}{\DeInferName\xspace}
\newcommand{\DeInferNoSpace}{\DeInferName}

\definecolor{BaseColor}{HTML}{FED976}
\definecolor{DeInferColor}{HTML}{D7F0D1}
\begin{document}

\title{\DeInfer: Efficient Parallel Inferencing for Decomposed Large Language Models}

\author{You-Liang Huang}
\email{yliangh@bu.edu}
\affiliation{%
  \institution{Boston University}
  \city{Boston}
  \state{Massachusetts}
  \country{USA}
}
\affiliation{%
  \institution{HKUST (Guangzhou)}
  \city{Guangzhou}
  \state{Guangdong}
  \country{China}
}

\author{Xinhao Huang}
\author{Chengxi Liao}
\email{xhuang171@connect.hkust-gz.edu.cn}
\email{cliao118@connect.hkust-gz.edu.cn}
\affiliation{%
  \institution{HKUST (Guangzhou)}
  \city{Guangzhou}
  \state{Guangdong}
  \country{China}
}

\author{Zeyi Wen}
\authornote{Corresponding Author}
\email{wenzeyi@hkust-gz.edu.cn}
\affiliation{%
  \institution{HKUST (Guangzhou) \& HKUST}
  \city{Guangzhou}
  \state{Guangdong}
  \country{China}
}

\renewcommand{\shortauthors}{Huang et al.}

\begin{abstract}
Existing works on large language model (LLM) decomposition mainly focus on improving performance on downstream tasks, but they ignore the poor parallel inference performance when trying to scale up the model size. To mitigate this important performance issue, this paper introduces \DeInfer, a high-performance inference system dedicated to parallel inference of decomposed LLMs. It consists of multiple optimizations to maximize performance and be compatible with state-of-the-art optimization techniques. Extensive experiments are carried out to evaluate \DeInfer's performance, where the results demonstrate its superiority, suggesting it can greatly facilitate the parallel inference of decomposed LLMs.
\end{abstract}




\maketitle

\section{INTRODUCTION}
\label{sec:intro}

\begin{figure}
\centering
\subfloat[Model under different compression ratios (8$\times$A800, 80GB)]{\includegraphics[width=\linewidth]{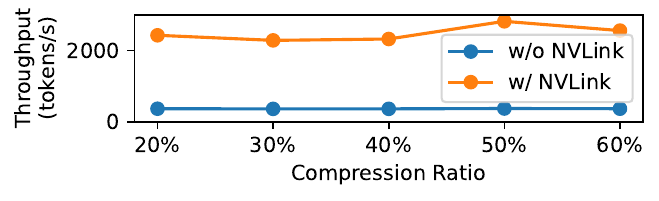}}
\vspace{-0.4em}
\subfloat[Model under different numbers of GPUs]{\includegraphics[width=\linewidth]{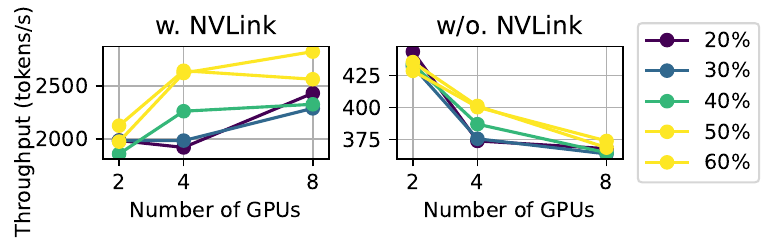}}
\vspace{-1em}
\caption{Parallel inference throughput of decomposed LLaMA-3-70B on a cluster of 8$\times$A800 (80GB).}
\label{fig:direct-scaling}
\vspace{-1em}
\end{figure}

Following the scaling law~\citep{ScalingLaws2020}, modern large language models (LLMs) usually have tens of billions of model parameters, which poses considerable challenges for model deployment and inference regarding memory footprint. To reduce the memory footprint, researchers have proposed a variety of model compression techniques, such as quantization~\citep{DuQuant2024}, model pruning~\citep{SparseGPT}, and model decomposition~\citep{SoLA,svd_llm}. Among these, decomposition-based compression has received relatively less attention than its counterparts. One key reason is that it is not inference-friendly, especially in a parallel setting.
The parallel inference performance of decomposed LLMs cannot scale with the increase of compression ratios, as shown in Fig.~\ref{fig:direct-scaling}(a), nor the parallelism, as shown in Fig.~\ref{fig:direct-scaling}(b).

To mitigate this performance issue, we propose \DeInfer, a high-performance inference system dedicated to the parallel inference of decomposed LLMs. In this paper, we first identify three key performance bottlenecks and then elaborate on the details of \DeInfer and how it is designed to eliminate these bottlenecks. After that, we carry out extensive and comprehensive experiments to evaluate the performance and scalability of \DeInfer. To the best of our knowledge, there is no other work focusing on improving the performance of decomposed LLM parallel inference. The code is available in \href{https://github.com/youliangh/DeInfer}{Github}.

% Planned experiment 1
% Scalability of other model compression methods

\vspace{-8pt}
\section{PRELIMINARIES OF LLM DECOMPOSITION}
\subsection{Background}

Low-rank decomposition can reduce model parameters by approximating the weight matrix with several smaller factorized matrices. Unlike hardware-dependent compression techniques (\eg unstructured pruning), low-rank decomposition is emerging as a promising technique for LLM compression \cite{fwsvd, asvd, svd_llm, SoLA, MoDeGPT, PaLU2025, eigen_attn}.
Taking truncated Singular Value Decomposition (SVD) as an example, it substitutes the weight matrix $W \in R^{m \times n}$ by the product of two smaller matrices $A \in R^{m \times k}$ and $B \in R^{k \times n}$ as follows.
\vspace{-0.5em}
\begin{equation}
 W \approx AB 
 \label{eq:eq2_1_1}
\end{equation}
where $A = (U_k \sqrt{\Sigma_k})$, $B = (\sqrt{\Sigma_k} V_k^T)$, $U_k \in R^{m \times k}$ and $V_k^T \in R^{k \times n}$ are the top-$k$ truncated matrices. $\sqrt{\Sigma_k} \in R^{k \times k}$ is a diagonal matrix by the square-roots of the corresponding top-$k$ singular values in $\Sigma$. This low-rank decomposition reduces the number of parameters from $m \times n$ to $(m + n) \times k$, where users balance the trade-off between performance and model size by adjusting $k$.

Model decomposition techniques have two advantages in terms of reducing the memory footprint. First, it can directly reduce the size of model parameters by replacing original pre-trained weights with two smaller matrices. Second, the KV cache can be compressed as a by-product of compressing $K$ and $V$ matrices in the attention layers, where the low-rank intermediate results between the two matrix multiplications now act as KV caches.

However, these advantages in terms of reducing the memory footprint can turn out to be drawbacks in terms of inference efficiency. Additional matrix multiplications brought about by low-rank matrices and KV cache reconstruction inevitably cause computational overhead. More importantly, we notice that it would incur more significant performance degradation in a parallel setting. In the next subsection, we will elaborate on performance bottlenecks and problems that we observed in practice.

\vspace{-8pt}
\subsection{Performance Bottlenecks}
\label{sec:bottleneck}

\subsubsection{Communication}
\label{sec:bottleneck-comm}
\textbf{In low-rank decomposition, approximating one weight matrix with two smaller matrices introduces more \textit{reduce-sum} operations in parallel inference.}

\begin{figure}[t]
\centering

\subfloat[Original attention layers]
{\includegraphics[width=\linewidth]{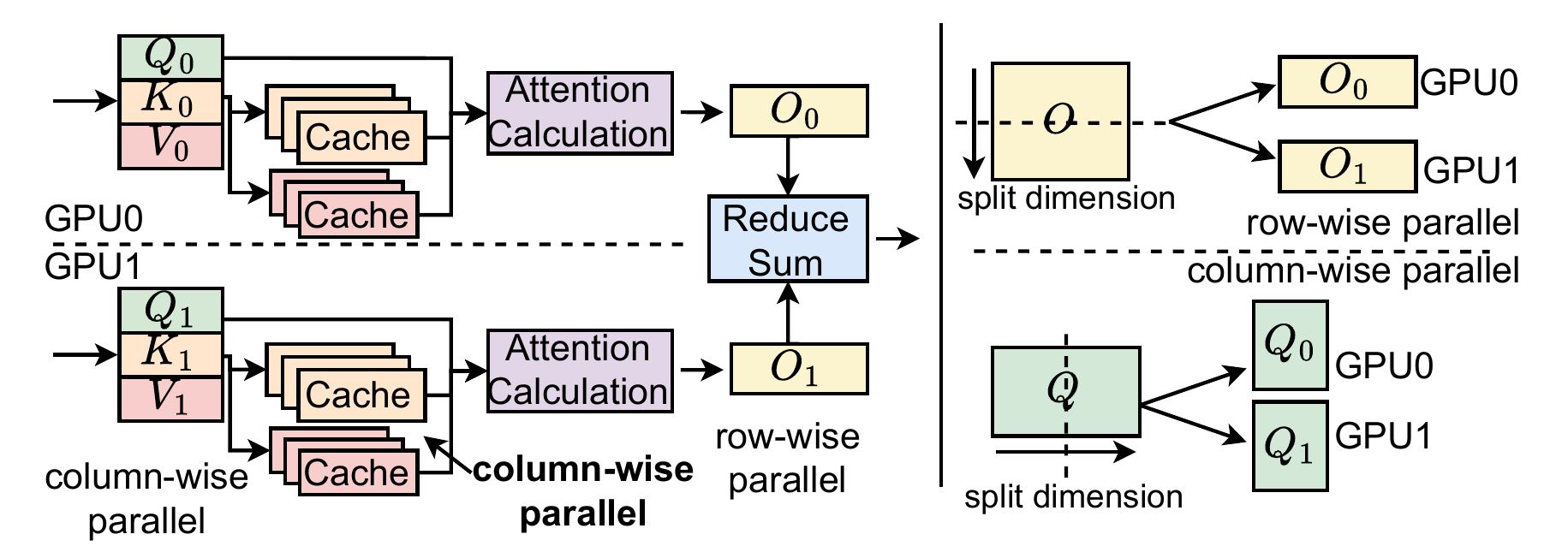}}

\subfloat[Decomposed attention layers]
{\includegraphics[width=\linewidth]{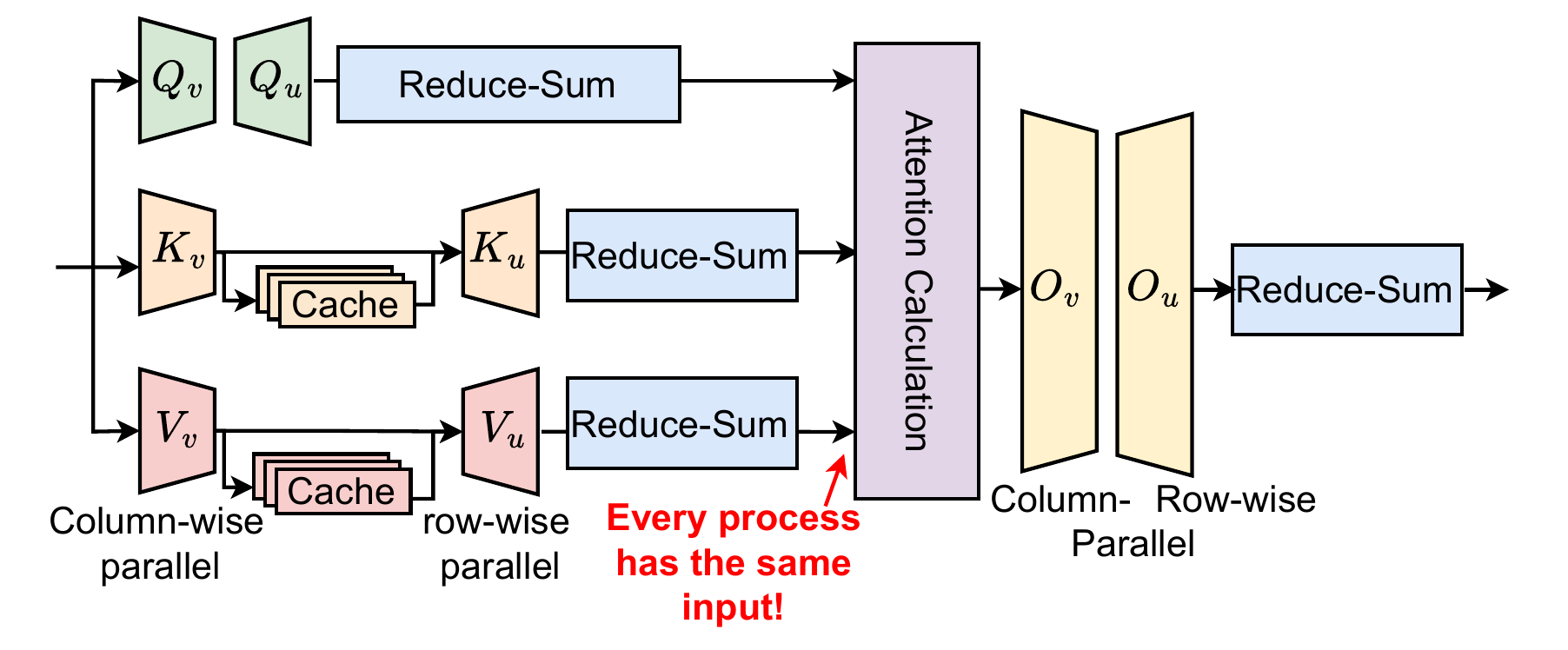}}

\vspace{-1em}
\caption{Self-attention computation is duplicated in parallel inference of decomposed LLMs.}
\label{fig:naive_attn_top}
\vspace{-1em}
\end{figure}

For example, in attention layers, as shown in Fig.~\ref{fig:naive_attn_top}(a), there are four matrices, \textit{Q}, \textit{K}, \textit{V}, and \textit{O} before the model decomposition. A common technique here for applying tensor parallelism is to split them into several shards. Every process holds a small chunk of matrices, where \textit{Q}, \textit{K}, and \textit{V} are column-wise splits, and \textit{O} is row-wise split. During the forward, there is only one \textit{reduce-sum} operation to accumulate the partial results of multiplication between \textit{O} and the result of self-attention. However, after model decomposition, as shown in Fig.~\ref{fig:naive_attn_top}(b), every matrix is replaced by two smaller matrices. For each paired low-rank matrices, there is a \textit{reduce-sum} operation to keep the computation correct. In comparison, there are four \textit{reduce-sum} operations in total instead of just one before model decomposition. More importantly, these additional operations are all \textit{reduce-sum}, the collective communication primitive that has the highest cost in terms of bandwidth and latency. The situation is the same for the MLP layers.

\vspace{-4pt}
\subsubsection{Duplicate Self-Attention Computation}
\textbf{In attention layers, self-attention computation cannot be parallelized.}

As depicted in Fig.~\ref{fig:naive_attn_top}(a), in the original model, with the column-wise parallel \textit{Q}, \textit{K}, and \textit{V}, every process has a small fraction of data and can perform self-attention computation on their own data. After multiplying the row-wise parallel \textit{O}, partial results will be accumulated with a \textit{reduce-sum} operation, and each process has an identical copy. However, after model decomposition, as shown in Fig.~\ref{fig:naive_attn_top}(b), there are three \textit{reduce-sum} operations after multiplying upward projection matrices. At this moment, every process has an identical copy of data. Therefore, each process would perform a duplicate self-attention computation, which is very costly and totally unnecessary.

\vspace{-4pt}
\subsubsection{Being incompatible with CUDA Graph} 
\textbf{CUDA Graph requires fixed arguments, but the results of the KV cache reconstruction have a dynamic shape.}

The latest LLM inference systems adopt paged KV cache management. Every process organizes a large chunk of GPU memory into a number of blocks, where a physical block can contain several slots of KV cache. Meanwhile, a block table is maintained to manage the mapping between logical blocks and physical blocks, where requests can only access logical blocks, and the inference system decides the swap-in and swap-out of physical blocks. When doing self-attention computation, the Flash Attention kernel takes the block table and base address as input, using an internal loop to fetch KV cache scatter in the blocks to perform the calculation~\citep{flash_attn_v1, flash_attn_v2}. For the original model, it is naturally compatible with a static computation graph, \ie CUDA Graph. However, for decomposed LLMs, things become totally different. During decoding, the number of KV cache is increasing, which means the size of KV cache reconstruction results are enlarging. The dynamic shape of the data conflicts with the requirement of building a static computation graph. Therefore, decomposed LLMs have to use a dynamic computation graph, which can cause considerable performance degradation.
\vspace{-6pt}

\section{RELATED WORKS}
\subsection{LLM Decomposition}
LLM decomposition adopts low-rank decomposition techniques that approximate weight matrices through products of low-rank matrices. Singular Value Decomposition (SVD) is a classic technique~\citep{fwsvd}, and the latest SVD-based methods introduce whitening techniques to mitigate the reconstruction error caused by outliers, which gives them superior performance in downstream tasks~\citep{asvd,svd_llm}. Atomic Feature Mimicking (AFM) is another common technique in LLM decomposition. It applies principle component analysis for the feature-based low-rank factorization~\citep{features_low_rank}. In addition to different matrix factorization, the search for better truncation positions now become another promising area for LLM decomposition~\citep{svd_llm2, bolaco, auto_trunc}. At the same time, researchers are also active in exploring the potential of low-rank decomposition in LLM inference. \citet{PaLU2025} and \citet{ShadowKV} tried to reduce memory footprint by storing cache in a low-rank manner. \citet{eigen_attn} operationalizes the attention mechanisms within low-dimensional eigenbases, achieving memory-efficient KV cache storage. Although much progress has been achieved in the development of LLM decomposition, there is still no attempt to explore what a decomposed LLM would be like in a parallel setting when using the state-of-the-art LLM inference systems (\eg vLLM~\citep{vLLM} and SGLang~\citep{zheng2024sglang}).

\vspace{-8pt}
\subsection{\DeInfer's Position}
We need to emphasize that the primary goal and position of our work is completely different from the LLM decomposition work we discussed above. \textit{Instead of providing a novel LLM decomposition technique that has better performance in downstream tasks, our work aims to improve the parallel inference performance of decomposed LLMs.} As for KV cache compression, storing the KV cache in a low-rank manner is a by-product of model decomposition and its reconstruction (on single GPU) is well-established in existing works. However, no attempts have been made to address the problems of KV cache reconstruction in a parallel setting (\S\ref{sec:bottleneck}), and our \DeInfer can address these problems.

\section{PROPOSED WORK}

In this section, our proposed system \DeInfer will be elaborated. First, we will introduce a novel low-rank communication technique that can significantly reduce communication costs in decomposed LLMs. Then, we will give the details of other optimizations. In the end, we will demonstrate how our \DeInfer can accommodate different LLM variants and model decomposition variants.

\vspace{-8pt}
\subsection{Highly Efficient Low-rank Communication}
\label{sec:design-low-rank-comm}
\begin{figure}[]
\vspace{-2em}
\centering

\subfloat[Attention layers]{\includegraphics[width=\linewidth]{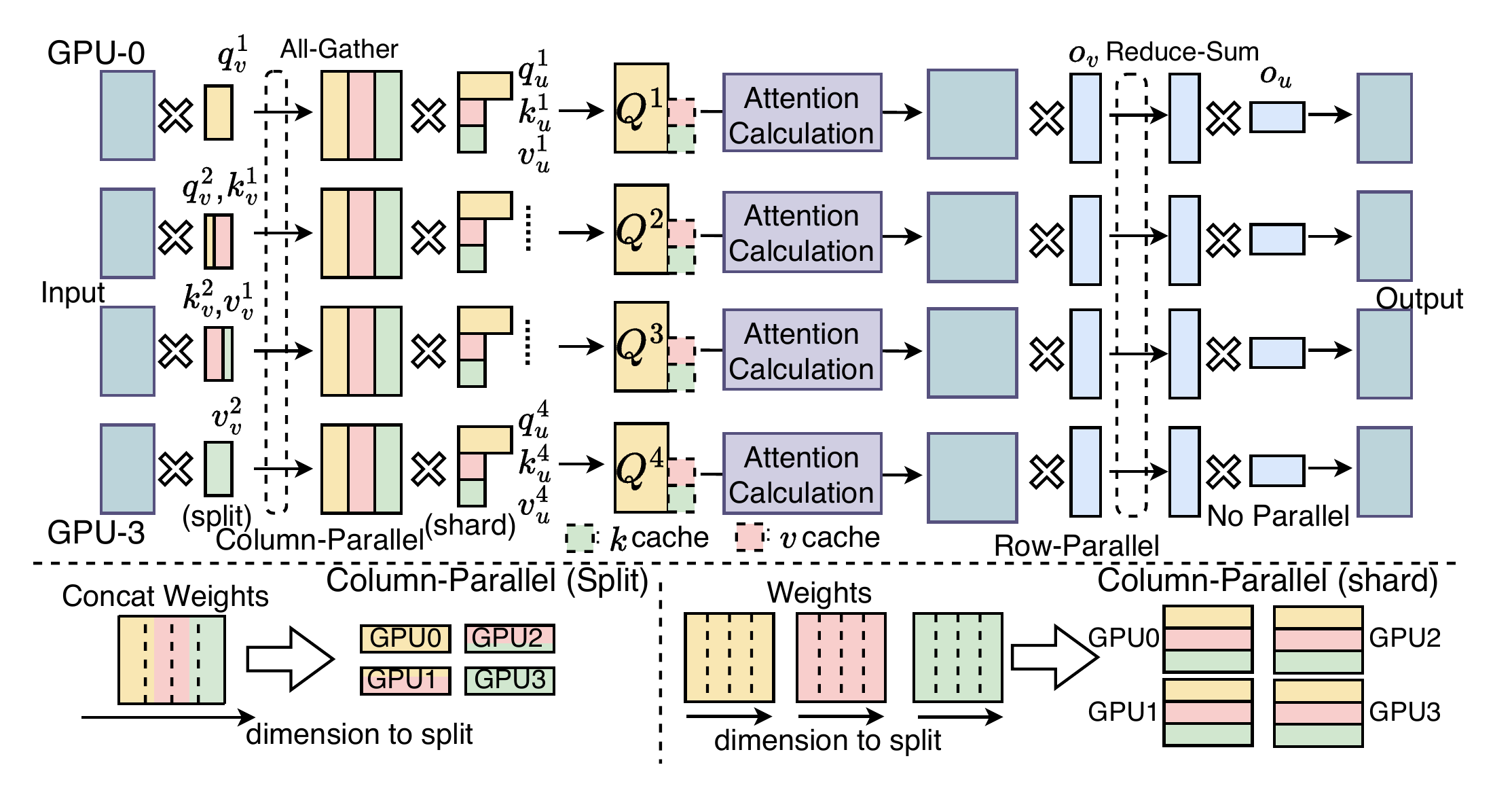}}

\subfloat[MLP layers]{\includegraphics[width=\linewidth]{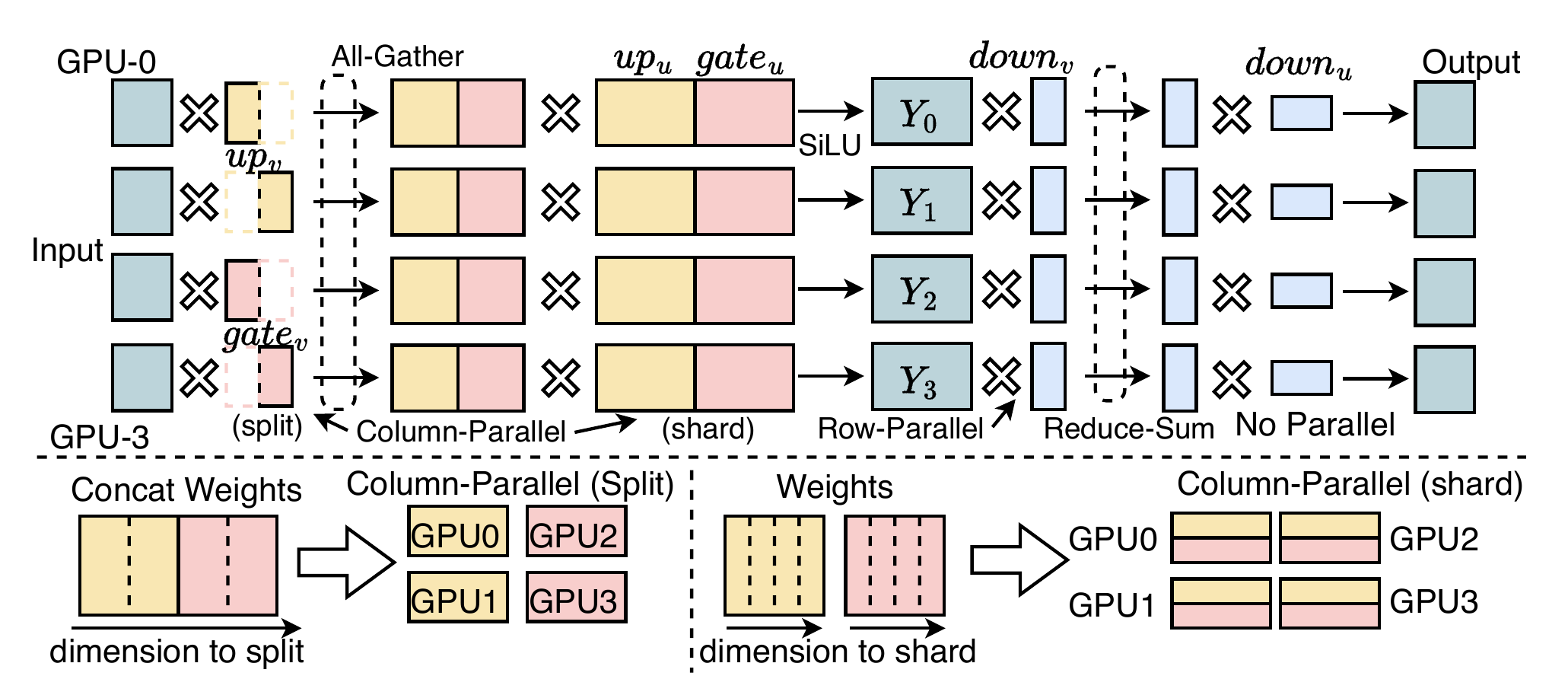}}
\vspace{-8pt}
\caption{Low-rank communication design in decomposed LLaMA models.}
\vspace{-1em}
\label{fig:low-rank-comm}
\end{figure}

After model decomposition, we can obtain a downward projection matrix $x_v$ and an upward projection matrix $x_u$ from the original weight matrix $x$. We noticed that most attention layers and MLP layers in modern LLMs both are two sub-layers, thus there would be four sub-layers in total after model decomposition. Our key idea is that, \textit{instead of having a reduce-sum operation every two sub-layers in the normal latent space, we can rearrange the current computation pipeline to let communication happen in the low rank latent space}. The rearranged computation pipelines are depicted in Fig.~\ref{fig:low-rank-comm}.

As shown in Fig.~\ref{fig:low-rank-comm}, all downward projection matrices $x_v$ in the first sub-layer are in column-wise parallel (in a \text{split} manner, where all matrices are first concatenated and then evenly split), followed by an \textit{all-gather} operation in the low-rank latent space. After \textit{all-gather}, each process now has identical low-rank data. In attention layers, the low rank data for \textit{K} and \textit{V} (\ie $k_{low\_rank}$ and $v_{low\_rank}$) will be stored as KV cache, whose reconstruction will be elaborated in the next subsection. The upward projection matrices $x_u$ are also in column-parallel (but in a \textit{shard} way, where each process has a shard of the same matrices). At this point, to keep correct computation, the forward performs a batched matrix multiplication for the low-rank results with different upward matrix shards $x_u$, respectively, \eg $k_{low\_rank}$ with $k_u$, and $v_{low\_rank}$ with $v_u$. Each process has their small chunk of data ready to proceed to perform self-attention or activation computation. \textit{It's noted that the decomposed LLMs now no longer have duplicate self-attention computation.} In the second sub-layer, the downward projection matrix is in row-wise parallel, and upward projection matrix is identical in each process. Therefore, after multiplying the downward matrix, each process needs to have a \textit{reduce-sum} operation to accumulate the partial results of all processes. Since every process has identical low-rank data and identical upward matrix, the output would also be identical.

\begin{table}[hb]
\centering
\resizebox{\linewidth}{!}{
\begin{threeparttable}
\begin{tabular}{|c|ccc|c|}
\hline
\multirow{2}{*}{Layer} & \multirow{2}{*}{Status} & \multicolumn{2}{c|}{Cost} & \multirow{2}{*}{\shortstack{Total\\Bandwidth}} \\
& & \textit{all-gather}, $O(n)$ & \textit{reduce-sum}, $2O(n)$ &  \\
\hline
\multirow{2}{*}{\shortstack{Attention}} & Unoptimized & 0 & $2\times(2h + 2h_{kv})$ &  $4h + 4h_{kv}$ \\
 & \DeInfer & $l_q+l_k+l_v$ & $2l_o$ & $l_q+l_k+l_v+2l_o$ \\
\hline
\multirow{2}{*}{\shortstack{MLP}} & Unoptimized & 0 & $2\times(2m+h)$ & $4m+2h$ \\
 & \DeInfer & $l_{up}+l_{gate}$ & $2l_{down}$ & $l_{up}+l_{gate}+2l_{down}$ \\
\hline
\multirow{2}{*}{\shortstack{LLaMA-3-70B}$^\dagger$} & Unoptimized & 0 & 167,936 & 167,936 \Original{100\%} \\
 & \DeInfer & 15,976 & $2\times9832$ & 35,640 \Improve{78\%} \\
\hline
\end{tabular}
\begin{tablenotes}
\item[$\dagger$]: In LLaMA-3-70B, we have $h=8192, h_{kv}=1024, m=28672$. Under the compression ratio of 40\% (evenly), $l_{up}=l_{down}=l_{gate}=l_{q}=l_{o}=60\%h=4916$ and $l_{k}=l_{v}=60\%h_{kv}=614$.
\end{tablenotes}
\end{threeparttable}
}
\caption{Communication cost per token in a single Transformer block during forward pass (LLaMA-3-70B).} % Amount of tensor per token in a forward pass
\label{tab:comm_cost}
\vspace{-1.5em}
\end{table}

To better understand how much bandwidth we can save through the communication, we analyze the bandwidth usage of LLaMA-3-70B in Table.~\ref{tab:comm_cost}, where we compare the unoptimized decomposed LLMs with our \DeInfer under the compression ratio of 40\%.

In Table.~\ref{tab:comm_cost}, $l_X$ means the number of reduced dimension of the matrix $X$, $h$ is the hidden dimension, $h_{kv}$ is the dimension of the \textit{K} and \textit{V} matrices, and $m$ is the intermediate dimension in MLP. And the bandwidth of \textit{reduce-sum} is double of \textit{all-gather}. As demonstrated, \DeInfer significantly reduces bandwidth usage by 78\% through rearranging the computation pipeline.

\vspace{-8pt}
\subsection{Integrating Paged Cache and CUDA Graph}
\begin{figure}[t]
\centering

\subfloat[Before model decomposition]{\includegraphics[width=\linewidth]{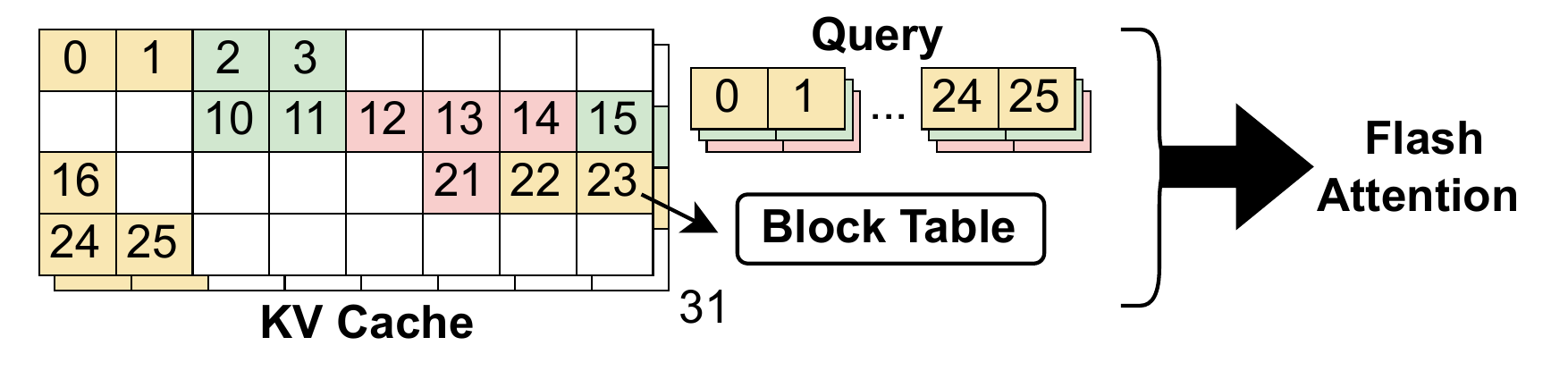}}

\subfloat[After model decomposition (\DeInfer)]
{\includegraphics[width=0.95\linewidth]{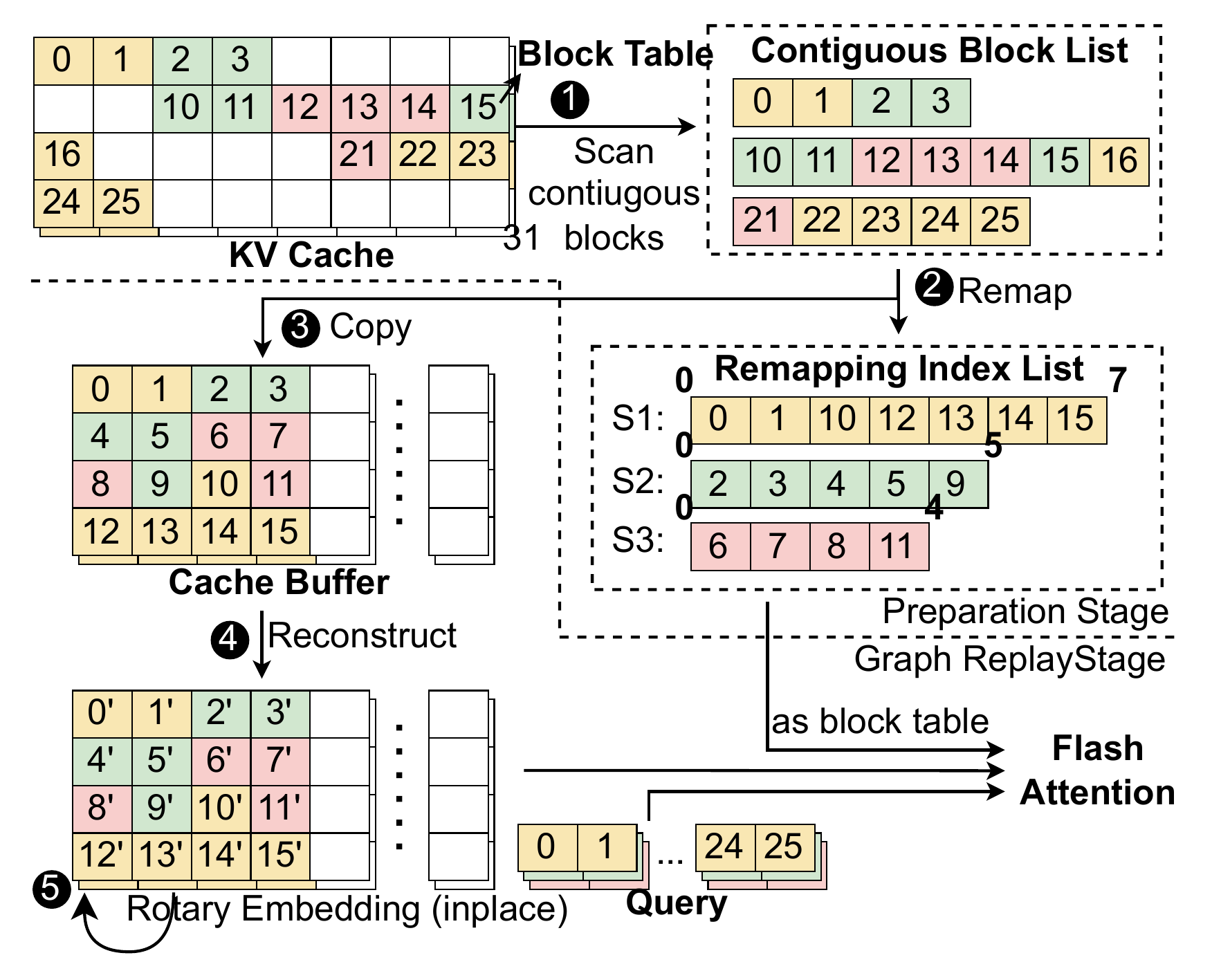}}
\vspace{-10pt}
\caption{KV cache reconstruction process.}
\label{fig:cache_copy}
\vspace{-0.5em}
\end{figure}

As depicted in Fig.~\ref{fig:cache_copy}(a), Flash Attention kernel takes the KV cache (\ie base address of KV cache area), \textit{Query} (\ie results of multiplying the \textit{Q} matrix), and a block table as input. All input is fixed during the forward. However, after model decomposition, decoding can no longer use the static computation graph since the shape of KV cache reconstruction results is dynamic. Therefore, we re-design the computation of KV cache reconstruction to allow the decoding to be compatible with both paged cache and CUDA Graph. Our designed process for KV cache construction during decoding is shown in Fig.~\ref{fig:cache_copy}(b). It has two stages, the \textit{preparation stage}, where we are allowed to do any CPU and GPU operations for the CUDA Graph execution, and the \textit{graph replay stage}, where only captured operations can be executed, and all operations must fulfill the requirements of CUDA Graph.

\DeInfer introduces two buffers for squeezing and temporally storing the KV cache and their corresponding reconstruction results. These two buffers are allocated in the initialization of the inference system. We noticed that iteratively copying the scattered KV cache is extremely time-consuming, thus \DeInfer tries to copy as many blocks as possible at one time. In the preparation stage, \DeInfer performs a scan to find which logical block is physically contiguous. At the same time, \DeInfer derives a new block table called \textit{remapping index list}. It acts as a block table, indicating which blocks in the buffer a sequence owns. In the graph replay stage, the physically contiguous KV caches are sequentially copied to the buffer, where the KV cache in the buffer is compact. Then, the compact KV cache multiplies upward matrices to finish the reconstruction. It is noted here that the results have a dynamic shape but a fixed memory address and we only need to use the GEMM kernel that has a large size to adapt for CUDA Graph. Overhead here is inevitable, but it can bring larger overall efficiency improvement by eliminating all kernel launch overhead. At this point, we can apply our customized in-place rotary position embedding kernel to the reconstruction results if the model needs. When everything is done, the buffer act as the original KV cache and \textit{remapping index list} as the block table. They and \textit{Query} are used to call Flash Attention kernel to complete self-attention computation.

In addition to the highly efficient customized kernel, \DeInfer also adopts other optimization techniques (\eg the communication-computation overlap and kernel fusion) to further improve the performance. We cannot introduce the technical details here due to page limit, but they can be found in \DeInfer's implementation.

\vspace{-10pt}
\subsection{Generalization}

Revisiting the low-rank communication in \S\ref{sec:design-low-rank-comm}, our proposed method does not place any constraint on the shape of the weight matrix. \DeInfer supports all tensor parallelism configurations as long as one matrix can be appropriately partitioned and evenly distributed. Therefore, \DeInfer can support decomposed LLMs with variadic compression ratios, \eg weights in different layers have different compression ratios, and even \textit{Q}, \textit{K} and \textit{V} in the same attention layer have different compression ratios. Additionally, for varying model decomposition methods, we have observed that there is a fundamental principle in matrix factorization, \ie no non-linear operations are introduced during factorization. \textit{This principle indicates that any more than two matrices derived from factorization can be absorbed into two matrices and then fit our \DeInfer.}

\begin{table}[h]
\centering
\resizebox{\linewidth}{!}{
\begin{threeparttable}
\begin{tabular}{|cccc|cc|c|}
\hline
\multicolumn{4}{|c|}{Attn} & \multicolumn{2}{|c|}{MLP} & \multirow{2}{*}{\shortstack{Rotary Position\\Embed}} \\
MHA & MQA & GQA & MLA$^\dagger$ & non-GLU & GLU-based & \\
\hline
$\checkmark$ & $\checkmark$ & $\checkmark$ &  & $\checkmark$ & $\checkmark$ & $\checkmark$ \\
\hline
\end{tabular}
\begin{tablenotes}
\item[$\dagger$]: Almost only be used in Deepseek's MoE models.
\end{tablenotes}
\end{threeparttable}
}
\caption{Attention and MLP variants supported by \DeInfer.}
\label{tab:generalization}
\vspace{-1.5em}
\end{table}

As for transformer-based LLMs, even though there are MoE decomposition methods~\citep{MoE-SVD}, our \DeInfer now only supports non-MoE models. The supported LLM variants are reported in Table~\ref{tab:generalization}. As shown, most variants can be supported by \DeInfer, which includes but is not limited to \textit{LLaMA}~\citep{LLaMA3}, \textit{OPT}~\citep{OPT}, and \textit{Qwen}~\citep{qwen3} models.

\vspace{-0.5em}
\section{EXPERIMENTS}
\label{sec:experiments}

In this section, we evaluate our proposed \DeInfer with the focus on the following three aspects: \textit{throughput analysis}, \textit{latency analysis}, and \textit{system-level analysis}. Additionally, we conduct experiments to demonstrate the necessity of supporting CUDA Graph and how is the scalability of \DeInfer. For demonstration, we implement \DeInfer based on one of the state-of-the-art LLM inferencing systems, \textit{vLLM}.

\vspace{-8pt}
\subsection{Setup}

Our experiments are conducted on two hardware platforms: (1) 8$\times$A800 (80GB) with 2$\times$Intel Xeon-8358P and 2TB memory. (2) 8$\times$A6000 (48GB) with 2$\times$Intel Xeon-8358 and 512GB memory, where platform (1) has fully-connected NVLinks but platform (2) doesn't have NVLinks. To demonstrate generalizability, we select LLaMA-65B, LLaMA-3-70B, and OPT-30B as foundation models because they encompass MHA and GQA in attention variants, non-GLU and GLU-based MLPs, and rotary embedding. The test scripts that we adopt are from the vLLM benchmark suite\footnote{https://github.com/vllm-project/vllm/tree/main/benchmarks} for fair and replicable comparison. \textit{To the best of our knowledge, there is no other work on parallel inference of decomposed LLMs. Therefore, we implement a basic implementation of tensor parallelism (hereinafter referred to as \textit{Base}).} Our experiments compare the performance of \DeInfer and Base, which also include system-level ablation studies and detailed analysis. Due to inconsistent GPU memory sizes between test platforms, we only keep configurations consistent in each experiment.

\vspace{-8pt}
\subsection{Throughput Analysis}
\label{sec:throughput}
\begin{figure}[ht]
\centering
\begin{subfigure}{\linewidth}
\includegraphics[width=\linewidth]{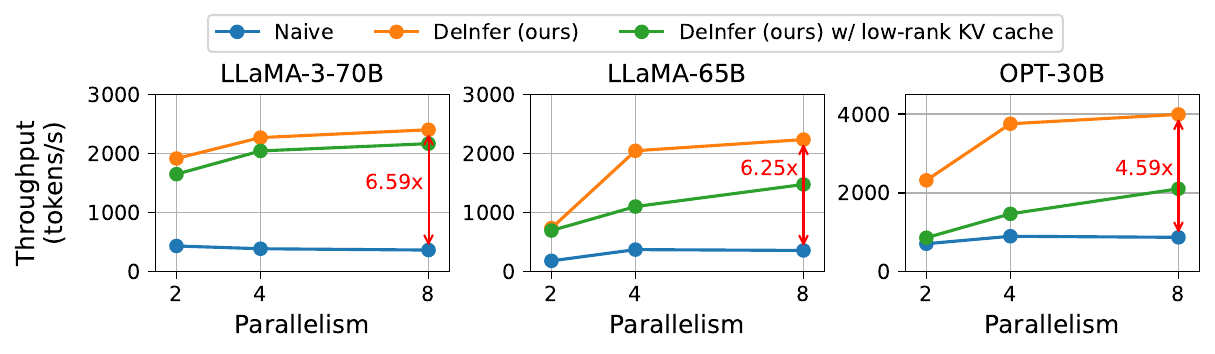}
\caption{Without NVLink}
\end{subfigure}
\begin{subfigure}{\linewidth}
\includegraphics[width=\linewidth]{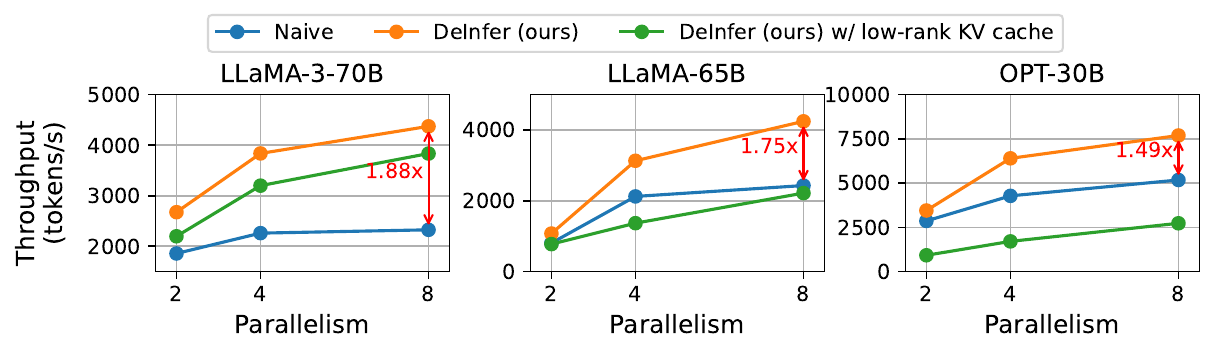}
\caption{With NVLink}
\end{subfigure}
\vspace{-20pt}
\caption{Batched generation performance of different models under 40\% compression ratio in the high load scenario on 8$\times$A800 (80GB).}
\label{fig:offline-throughput-s1}
\end{figure}

\begin{figure}[h]
\centering
\begin{subfigure}{\linewidth}
\includegraphics[width=\linewidth]{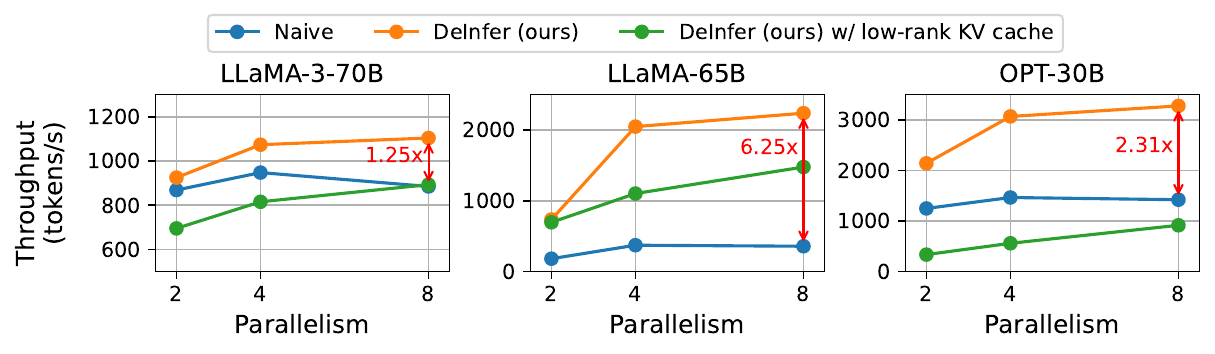}
\caption{Without NVLink}
\end{subfigure}
\begin{subfigure}{\linewidth}
\includegraphics[width=\linewidth]{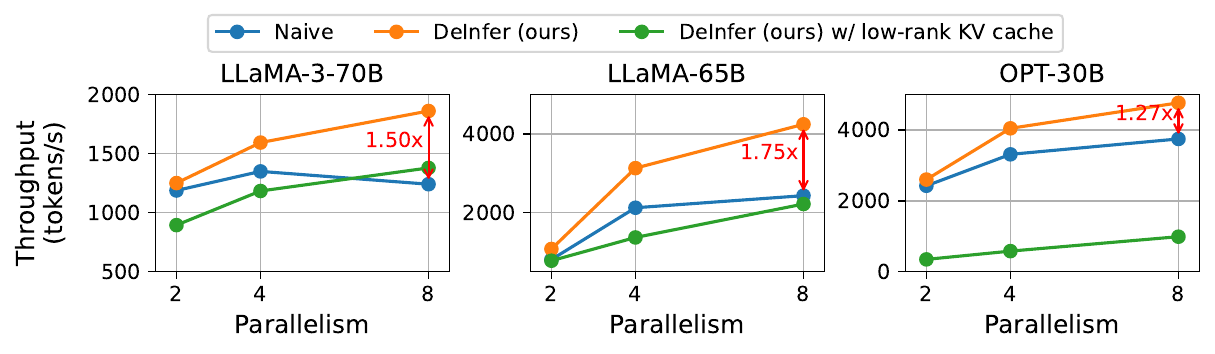}
\caption{With NVLink}
\vspace{-1em}
\end{subfigure}
\caption{Batched generation performance of different models in the mild load scenario under 40\% compression ratio on 8$\times$A800 (80GB).}
\label{fig:offline-throughput-s2}
\vspace{-0.1em}
\end{figure}

We prepare two different scenarios to evaluate \DeInfer's throughput performance. In the first setting, there is a large batch size (512 for LLaMA-3-70B and LLaMA-65B, 768 for OPT-30B) but with a relatively smaller maximal sequence length (32 pre-fill tokens and the maximal sequence length is 160), whereas in the second setting, it is vice versa. The batch size is set to 64 for LLaMA-3-70B and LLaMA-65B, 192 for OPT-30B, where pre-fill tokens is 64 and the maximal sequence length is 512. The configuration is kept the same for each experiment setup. The results are shown in Fig.~\ref{fig:offline-throughput-s1} and Fig.~\ref{fig:offline-throughput-s2}, respectively for the first and the second setting. For the both scenarios, as shown in the figures, Base has poor performance and scalability on the three different models. In comparison, \DeInfer shows a much better scalability and significantly outperforms Base by different extents.

%              w/o.   w/  NVLink
% LLaMA-3-70B: 6.60x, 1.88x
% LLaMA-65B  : 6.25x, 1.75x
% OPT-30B    : 4.59x, 1.49x

% Additionally, we have noticed that, when NVLink is not enabled, as the batch size decreases, the performance of the Base implementation improves, whereas \DeInfer's performance deteriorates, indicating that the major performance bottleneck here is the communication. A smaller batch size for \DeInfer underuses bandwidth, and a larger batch size overwhelms Base. A detailed analysis is elaborated in \S\ref{sec:sys-profile}.

%              w/o.   w/  NVLink
% LLaMA-3-70B: 1.25x, 1.5x
% LLaMA-65B  : 6.25x, 1.75x
% OPT-30B    : 2.31x, 1.27x

\vspace{-1em}
Additionally, when enabling low-rank KV cache, there is a noticeable performance gap between the GQA model (LLaMA-3-70B) and the MHA models (LLaMA-65B and OPT-30B). This is because MHA models have more attention heads (64 in LLaMA-65B vs. 8 in LLaMA-3-70B), which means higher KV cache reconstruction cost. This difference brings tremendous advantages for LLaMA-3-70B in rebuilding the KV cache. In comparison, LLaMA-65B suffers a much greater performance degradation when enabling low-rank KV cache. Similarly, the OPT-30B can also witness the same problem.

\vspace{-8pt}
\subsection{Latency Analysis}
In this subsection, we adopt the vLLM serving benchmark to evaluate the latency performance of our system, where there are three metrics used for performance assessment: Time-to-First Token (TTFT) and Inter-token Latency (ITL). The request in this experiment has 128 prompts that have 32 pre-fill tokens, and the system is required to generate 256 tokens for each prompt. The configuration is kept the same across different platforms. Experiments are conducted on the A800 platform and A6000 platform, whose results are reported in Table~\ref{tab:serving-a6000} and Table~\ref{tab:serving-a800}, respectively.

\begin{table}[h]
\resizebox{\linewidth}{!}{
\begin{threeparttable}
\begin{tabular}{|c|c|r|r|r|}
\hline
\multirow{2}{*}{Metrics} & \multirow{2}{*}{Methods} & \multicolumn{1}{c|}{\multirow{2}{*}{LLaMA-3-70B}} & \multicolumn{1}{c|}{\multirow{2}{*}{LLaMA-65B}} & \multicolumn{1}{c|}{\multirow{2}{*}{OPT-30B}} \\
& && & \\ \hline
\multirow{3}{*}{TTFT (ms)} & Base & 21740& 19341 & 6999 \\
& \DeInfer & \Improve{83\%} 3633& \Improve{82\%} 3489 & \Improve{77\%} 1594 \\
& \DeInferNoSpace$^\dagger$ & \Improve{83\%} 3703& \Improve{81\%} 3626 & \Improve{76\%} 1704 \\
\hline
\multirow{3}{*}{ITL (ms)}& Base & 812 & 764 & 311\\
& \DeInfer & \Improve{79\%} 172 & \Improve{76\%} 181 & \Improve{68\%} 101\\
& \DeInferNoSpace$^\dagger$ & \Improve{74\%} 211 & \Improve{59\%} 312 & \Improve{39\%} 190 \\ \hline
\end{tabular}
\begin{tablenotes}
\item[$\dagger$]: w/ enabling low-rank KV cache.
\end{tablenotes}
\end{threeparttable}
}
\caption{Serving performance of different models under a compression ratio of 40\% on 8xA6000 (48GB) w/o. NVLink}
\label{tab:serving-a6000}
\vspace{-2em}
\end{table}

\begin{table*}[ht]
\resizebox{\textwidth}{!}{
\begin{threeparttable}
\begin{tabular}{|r|l|rrr|rrr|rrr|}
\hline
\multicolumn{11}{|c|}{\textbf{w/o. NVLink}} \\
\hline
\multicolumn{1}{|c|}{\multirow{2}{*}{Metrics}} & \multirow{2}{*}{Methods} & \multicolumn{3}{c|}{LLaMA-3-70B} & \multicolumn{3}{c|}{LLaMA-65B}& \multicolumn{3}{c|}{OPT-30B}\\ \cline{3-11} 
& & \multicolumn{1}{c}{TP=2} & \multicolumn{1}{c}{TP=4} & \multicolumn{1}{c|}{TP=8} & \multicolumn{1}{c}{TP=2} & \multicolumn{1}{c}{TP=4} & \multicolumn{1}{c|}{TP=8} & \multicolumn{1}{c}{TP=2} & \multicolumn{1}{c}{TP=4} & \multicolumn{1}{c|}{TP=8} \\ \hline
\multirow{3}{*}{TTFT (ms)} & Base & 8950& 10064 & 10764 & 8164& 9359& 10451 & 3126& 3566 & 3845 \\
& \DeInfer & \Improve{76\%} 2137 & \Improve{79\%} 2110 & \Improve{79\%} 2246 & \Improve{73\%} 2192 & \Improve{78\%} 2088 & \Improve{83\%} 1812 & \Improve{69\%} 960 & \Improve{74\%} 927& \Improve{75\%} 944\\
& \DeInferNoSpace$^\dagger$ & \Improve{72\%} 2484 & \Improve{79\%} 2102 & \Improve{81\%} 2093 & \Improve{73\%} 2216 & \Improve{80\%} 1840 & \Improve{80\%} 2085 & \Improve{70\%} 940 & \Improve{75\%} 878 & \Improve{76\%} 906 \\ 
\hline
\multirow{3}{*}{ITL (ms)}& Base & 156 & 122 & 96 & 206 & 103 & 101 & 54 & 47 & 47 \\
& \DeInfer & \Improve{31\%} 108 & \Improve{28\%} 88 & \Improve{9\%} 87 & \Improve{44\%} 115 & \Improve{12\%} 91 & \Improve{13\%} 88 & \Decrease{6\%} 57 & \Improve{0\%} 47 & \Improve{2\%} 46
\\
& \DeInferNoSpace$^\dagger$ & 
\Improve{8\%} 143 & \Improve{6\%} 115 & \Decrease{9\%} 105 & \Decrease{220\%} 660 & \Decrease{244\%} 354 & \Decrease{114\%} 216 & \Decrease{424\%} 283 & \Decrease{228\%} 154 & \Decrease{115\%} 101
\\ 
\hline
\hline
\multicolumn{11}{|c|}{\textbf{w/ NVLink}} \\
\hline
\multicolumn{1}{|c|}{\multirow{2}{*}{Metrics}} & \multirow{2}{*}{Methods} & \multicolumn{3}{c|}{LLaMA-3-70B} & \multicolumn{3}{c|}{LLaMA-65B}& \multicolumn{3}{c|}{OPT-30B}\\ \cline{3-11} 
& & \multicolumn{1}{c}{TP=2} & \multicolumn{1}{c}{TP=4} & \multicolumn{1}{c|}{TP=8} & \multicolumn{1}{c}{TP=2} & \multicolumn{1}{c}{TP=4} & \multicolumn{1}{c|}{TP=8} & \multicolumn{1}{c}{TP=2} & \multicolumn{1}{c}{TP=4} & \multicolumn{1}{c|}{TP=8} \\ \hline
\multirow{3}{*}{TTFT (ms)} & Base & 1662& 1499 & 1605 & 1525& 1417& 1461 & 724& 654 & 649 \\
& \DeInfer & \Improve{12\%} 1463 & \Improve{34\%} 983 & \Improve{50\%} 797 & \Improve{24\%} 1162 & \Improve{38\%} 875 & \Improve{49\%} 749 & \Improve{20\%} 581 & \Improve{32\%} 448 & \Improve{32\%} 443\\
& \DeInferNoSpace$^\dagger$ & \Improve{17\%} 1377 & \Improve{37\%} 948 & \Improve{49\%} 817 & \Improve{20\%} 1214 & \Improve{36\%} 901 & \Improve{46\%} 782 & \Improve{12\%} 640 & \Improve{21\%} 514 & \Improve{30\%} 457 \\ 
\hline
\multirow{3}{*}{ITL (ms)}& Base & 156 & 115 & 93 & 191 & 98 & 94 & 50 & 42 & 41 \\
& \DeInfer & \Improve{42\%} 90 & \Improve{34\%} 76 & \Improve{16\%} 78 & \Improve{52\%} 92 & \Improve{35\%} 64 & \Improve{40\%} 56 & \Improve{0\%} 50 & \Improve{12\%} 37 & \Improve{20\%} 33 \\
& \DeInferNoSpace$^\dagger$ & \Improve{19\%} 127 & \Improve{17\%} 96 & \Improve{0\%} 93 & \Decrease{236\%} 642 & \Decrease{237\%} 330 & \Decrease{96\%} 184 & \Decrease{452\%} 276 & \Decrease{248\%} 146 & \Decrease{110\%} 86 \\ \hline
\end{tabular}
\begin{tablenotes}
\item[$\dagger$]: w/ enabling low-rank KV cache.
\end{tablenotes}
\end{threeparttable}
}
\caption{Serving performance of different models under compression ratio of 40\% on A800 (80GB) platform}
\label{tab:serving-a800}
\vspace{-25pt}
\end{table*}

As reported, our \DeInfer achieves significant latency decreases in both test platforms and different NVLink settings. When enabling low-rank KV cache, the performance of \DeInfer shows different patterns. Generally, \DeInfer suffers more in the MHA models (\ie LLaMA-65B and OPT-30B) than the GQA model (LLaMA-3-70B). The reason remains the same: more attention heads in MHA models, which can also explain why enabling NVLink does not help. However, \DeInfer on the A6000 platform (Table~\ref{tab:serving-a6000}) shows a different performance pattern compared to it on the A800 platform, where on 8xA6000 we can observe a significant performance improvement when enabling low-rank KV cache. To explore the underlying reason, we profile the execution of Base and \DeInfer, and the details are elaborated in the next subsection. \label{sec:serving-discussion}

\vspace{-8pt}
\subsection{System Profiling}
\label{sec:sys-profile}
We report the breakdown of computation and communication profiled by NVIDIA Nsight Systems\footnote{https://developer.nvidia.com/nsight-systems} in Table~\ref{tab:breakdown}. As shown, Base's communication costs are predominant in experiments that do not enable NVLink, and things get worse as the parallelism increases. The exorbitant communication cost is thus the root cause of poor parallel inference performance. In comparison, \DeInfer improves communication efficiency by reducing 80\textasciitilde90\% communication cost. As parallelism increases, \DeInfer shows incredible scalability. Moreover, since \DeInfer eliminates duplicate self-attention computation, the computing time is reduced by 10\textasciitilde30\%.

\begin{table}[ht]
\centering
\resizebox{\linewidth}{!}{
\begin{threeparttable}
\begin{tabular}{|c|rrr|rrr|}
\hline
\multirow{3}{*}{TP} & \multicolumn{3}{c|}{\multirow{2}{*}{w/o. NVLink}} & \multicolumn{3}{c|}{\multirow{2}{*}{w/ NVLink}} \\
 & \multicolumn{3}{l|}{} & \multicolumn{3}{l|}{} \\ \cline{2-7}
 & compute (ms) & comm. (ms) & total (ms) & compute (ms) & comm. (ms) & total (ms) \\ \hline
\multirow{2}{*}{2} & \cellcolor{BaseColor}\Original{16\%} 2.352 & \cellcolor{BaseColor}\Original{84\%} 12.182 & \cellcolor{BaseColor}\Original{100\%} 14.534 & \cellcolor{BaseColor}\Original{73\%} 2.391 & \cellcolor{BaseColor}\Original{27\%} 0.891 & \cellcolor{BaseColor}\Original{100\%} 3.282 \\ \cline{2-7}
 & \cellcolor{DeInferColor}\Improve{23\%} 1.809 & \cellcolor{DeInferColor}\Improve{92\%} 0.977 & \cellcolor{DeInferColor}\Improve{81\%} 2.786 & \cellcolor{DeInferColor}\Improve{25\%} 1.792 & \cellcolor{DeInferColor}\Improve{74\%} 0.232 & \cellcolor{DeInferColor}\Improve{39\%} 2.014 \\ \hline
\multirow{2}{*}{4} & \cellcolor{BaseColor}\Original{7\%} 1.174 & \cellcolor{BaseColor}\Original{93\%} 14.853 & \cellcolor{BaseColor}\Original{100\%} 16.027 & \cellcolor{BaseColor}\Original{49\%} 1.176 & \cellcolor{BaseColor}\Original{51\%} 1.240 & \cellcolor{BaseColor}\Original{100\%} 2.416 \\ \cline{2-7}
 & \cellcolor{DeInferColor}\Improve{12\%} 1.038 & \cellcolor{DeInferColor}\Improve{92\%} 1.156 & \cellcolor{DeInferColor}\Improve{86\%} 2.194 & \cellcolor{DeInferColor}\Improve{11\%} 1.041 & \cellcolor{DeInferColor}\Improve{78\%} 0.267 & \cellcolor{DeInferColor}\Improve{46\%} 1.308 \\ \hline
\multirow{2}{*}{8} & \cellcolor{BaseColor}\Original{6\%} 1.112 & \cellcolor{BaseColor}\Original{94\%} 16.051 & \cellcolor{BaseColor}\Original{100\%} 17.163 & \cellcolor{BaseColor}\Original{45\%} 1.116 & \cellcolor{BaseColor}\Original{55\%} 1.369 & \cellcolor{BaseColor}\Original{100\%} 2.485 \\ \cline{2-7}
 & \cellcolor{DeInferColor}\Improve{29\%} 0.787 & \cellcolor{DeInferColor}\Improve{92\%} 1.285 & \cellcolor{DeInferColor}\Improve{88\%} 2.072 & \cellcolor{DeInferColor}\Improve{30\%} 0.782 & \cellcolor{DeInferColor}\Improve{82\%} 0.247 & \cellcolor{DeInferColor}\Improve{59\%} 1.029 \\ \hline
\end{tabular}
\begin{tablenotes}
\item Tag: \colorbox{BaseColor}{Base}, \colorbox{DeInferColor}{\DeInfer}. \Original{X\%} indicates the proportion of total cost. \Improve{X\%} denotes time saved by \DeInfer compared to Base.
\end{tablenotes}
\end{threeparttable}
}
\caption{Breakdown of execution time in a LLaMA-3-70B transformer layer on A800 with compression ratio of 40\%.}
\label{tab:breakdown}
\vspace{-2em}
\end{table}

According to hardware specifications, NVIDIA RTX A800 (80GB)\footnote{https://www.techpowerup.com/gpu-specs/a800-sxm4-80-gb.c3966} has a much larger memory bus and bandwidth than NVIDIA RTX A6000\footnote{https://www.techpowerup.com/gpu-specs/rtx-a6000.c3686}. As for the discussion at the end of the last subsection (\ie distinct performance on different platforms), inference performance on the A6000 platform is not only limited to exorbitant communication cost but also to much smaller memory bandwidth. The advantage provided by \DeInfer is so significant that it compensates more enough than the KV cache reconstruction cost. Therefore, we can witness a consistent performance improvement of \DeInfer in the three models of different architectures on the A6000 platform.

\vspace{-8pt}
\subsection{Scalability Analysis}
\begin{figure}[h]
\centering
\includegraphics[width=0.9\linewidth]{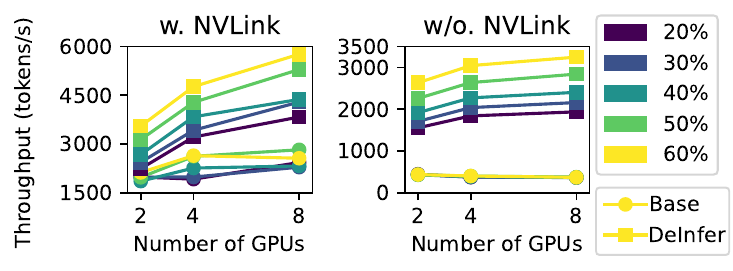}
\vspace{-1em}
\caption{Generation performance of LLaMA-3-70B of different compression ratios on A800 platform.}
\label{fig:scaling-compare}
\vspace{-7pt}
\end{figure}

As demonstrated in Introduction (\S\ref{sec:intro}), one of the most important motivations of our work is the poor scalability of decomposed LLMs, which fails to scale up the performance as the compression ratio or the parallelism increases. In this experiment, we run benchmarks (\ie the high-load generation experiment in \S\ref{sec:throughput}) on LLMs that are compressed to different extents to evaluate the scalability of \DeInfer. The results are shown in Fig.~\ref{fig:scaling-compare}, where \DeInfer shows excellent scalability. When the parallelism is 8 and compression ratio is 60\%, \DeInfer achieves a speed up of 2.25x and 8.81x for w/ NVLink and w/o. NVLink, respectively. Moreover, as the compression ratio increases, the performance of \DeInfer can also gradually improve.

\vspace{-8pt}
\subsection{Benefits of CUDA Graph}
Base's decoding stage cannot support the static graph, which may cause significant performance degradation. To explore how much performance can be improved, we test two types of \DeInfer, one with CUDA Graph and another without, in both throughput and serving experiments. The results are reported in Table~\ref{tab:cuda_graph_exp}.

\begin{table}[h]
\resizebox{\linewidth}{!}{
\begin{threeparttable}
\begin{tabular}{|c|c|r|r|r|r|}
\hline
\multirow{2}{*}{Experiments} & \multirow{2}{*}{Configurations} & \multicolumn{2}{c|}{LLaMA-3-70B} & \multicolumn{2}{c|}{LLaMA-65B} \\
& & \multicolumn{1}{c}{w/o. NVLink} & \multicolumn{1}{c|}{w/ NVLink} & \multicolumn{1}{c}{w/o NVLink} & \multicolumn{1}{c|}{w/ NVLink} \\ \hline
\multirow{4}{*}{Throughput} & \multirow{2}{*}{s1 (tokens/s)} & \cellcolor{BaseColor} 2068 & \cellcolor{BaseColor} 3488 & \cellcolor{BaseColor} 1435 & \cellcolor{BaseColor} 2011 \\
& & \cellcolor{DeInferColor}\ThroughputImprove{5\%} 2168 & \cellcolor{DeInferColor}\ThroughputImprove{10\%} 3834 & \cellcolor{DeInferColor}\ThroughputImprove{3\%} 1477 & \cellcolor{DeInferColor}\ThroughputImprove{9\%} 2217 \\ \cline{2-6} 
& \multirow{2}{*}{s2 (tokens/s)} & \cellcolor{BaseColor} 816 & \cellcolor{BaseColor} 831 & \cellcolor{BaseColor} 552 & \cellcolor{BaseColor} 660 \\
& & \cellcolor{DeInferColor}\ThroughputImprove{20\%} 978 & \cellcolor{DeInferColor}\ThroughputImprove{66\%} 1379 & \cellcolor{DeInferColor}\ThroughputImprove{6\%} 587 & \cellcolor{DeInferColor}\ThroughputImprove{8\%} 713 \\ \hline
\multirow{2}{*}{Serving} & \multirow{2}{*}{ITL (ms)} & \cellcolor{BaseColor} 118 & \cellcolor{BaseColor} 97 & \cellcolor{BaseColor} 279 & \cellcolor{BaseColor} 249 \\
& & \cellcolor{DeInferColor}\Improve{3\%} 114 & \cellcolor{DeInferColor}\Improve{5\%} 93 & \cellcolor{DeInferColor}\Improve{28\%} 202 & \cellcolor{DeInferColor}\Improve{31\%} 173 \\ \hline
\end{tabular}
\begin{tablenotes}
\small
\item Tag: \colorbox{BaseColor}{Base}, \colorbox{DeInferColor}{\DeInfer}. Table reports performance of \DeInfer that enables low-rank KV cache.
\end{tablenotes}
\end{threeparttable}
}
\normalsize
\caption{Performance improvement brought by static kernel execution graph, on 8xA800.}
\label{tab:cuda_graph_exp}
\vspace{-2em}
\end{table}

Table~\ref{tab:cuda_graph_exp} shows a general performance pattern. When NVLink is not enabled, the performance bottleneck remains the communication; therefore, supporting CUDA Graph does not help much. However, when NVLink is enabled and the performance bottleneck becomes the computation, we can notice a significant performance improvement in most scenarios. This is because the overhead of CUDA kernel launch is eliminated, which saves tremendous time and thus brings such a performance gain.
\vspace{-6pt}
\section{CONCLUSION}
\label{sec:conclusion}
To mitigate the performance issue of decomposed LLM parallel inference, this work proposes \DeInfer, a high-performance inference system dedicated to parallel inference of decomposed LLMs. It consists of a novel low-rank communication technique and other optimizations that can significantly improve the parallel inference performance. We integrate it into one of the state-of-the-art inference systems, \textit{vLLM}. Extensive experiments are carried out, where results demonstrate the excellent scalability and efficiency of \DeInfer, suggesting the practicality of \DeInfer in parallel inference of decomposed LLMs.

%% Bibliography
\bibliographystyle{ACM-Reference-Format}
\bibliography{references}

\end{document}